\documentclass[a4paper,conference]{IEEEtran}
\ifCLASSINFOpdf
\else
\fi
\usepackage{tikz}
\usetikzlibrary{decorations.pathreplacing}
\usetikzlibrary{fit}
\usetikzlibrary{decorations.markings,bending}
\usepackage{algorithm} 
\usepackage[noend]{algorithmic}
\usepackage{graphicx}
\usepackage{comment}
\usepackage{amsmath,amssymb} 
\usepackage{color}
\usepackage{relsize}

\newcommand\blfootnote[1]{%
  \begingroup
  \renewcommand\thefootnote{}\footnote{#1}%
  \addtocounter{footnote}{-1}%
  \endgroup
}

\begin{document}
%
\title{SSDL: Self-Supervised Domain Learning for Improved Face Recognition
}


\author{\IEEEauthorblockN{Samadhi~Wickrama Arachchilage and Ebroul Izquierdo}
\IEEEauthorblockA{Multimedia and Vision Group\\ School of Electronic Engineering and Computer Science\\
Queen Mary University of London, United Kingdom\\
Email: \{s.wickramaarachchilage, ebroul.izquierdo\}@qmul.ac.uk}
}


%


\maketitle

\begin{abstract}
Face  recognition  in  unconstrained  environments  is  challenging due  to variations in illumination, quality of sensing,  motion blur and etc. An individual’s face appearance can vary drastically under different conditions creating a gap between train (source) and varying test (target) data. The domain gap could cause decreased performance levels in direct knowledge transfer from source to target. Despite  fine-tuning with domain specific data could be an effective solution, collecting and  annotating  data  for  all domains  is extremely  expensive. To this end, we propose a self-supervised domain learning (SSDL) scheme that trains on triplets mined from unlabelled data. A key factor in effective discriminative  learning, is selecting  informative  triplets. Building on most confident predictions, we follow  an  “easy-to-hard”  scheme  of  alternate triplet mining and self-learning. Comprehensive experiments on four different benchmarks show that SSDL generalizes well on different domains. 

\end{abstract}


%
\IEEEpeerreviewmaketitle

\section{Introduction} 

\blfootnote{© 20XX IEEE.  Personal use of this material is permitted.  Permission from IEEE must be obtained for all other uses, in any current or future media, including reprinting/republishing this material for advertising or promotional purposes, creating new collective works, for resale or redistribution to servers or lists, or reuse of any copyrighted component of this work in other works.}

Face is an easy-to-extract biometric trait with high potential in practical application. Among others, it has shown excellent capabilities in security applications such as intelligent surveillance, user authentication applications such as traveler verification at border crossing points and diverse other mobile and social media applications. Consequently, a plethora of face recognition systems based on hand-crafted and lately deep Convolutional Neural Networks (CNNs) have been proposed and studied over the last few years. These systems have reported near-human and even advanced performance levels under controlled environments \cite{FaceNet,DeepIdTWOPLUS,DeepFace}. None-the-less, face recognition for practical application remains an open problem \cite{FaceSurvSAM, ijb_a}. Real-world applications have distinct settings  with varying levels of illuminations, camera quality and angle, motion levels, pose spectrum, biological specificities like ethnicity and etc. Hence, learning a universal representation for all possible settings is extremely challenging. 
\medskip

The main challenge in cross domain recognition lies in the inherent train data bias of deep networks.  Employing a network on a different domain requires minimizing the train data bias. Generic solutions like increasing training data or going deeper with convolutions, result in more powerful face descriptors, but do not necessarily undo the dataset bias. Since the impact of dataset bias fluctuates based on the distance between train and test domains, a more elegent solution is to generalize on different domains by learning and adapting. Yosinski et al. shows that the desired adapting could be achieved by fine-tuning the network on application specific data \cite{NIPS2014_5347}. However, the underlying process of data annotation  is time-consuming and complicated when the process has to be repeated on each new domain. Focusing on this challenging problem of domain invariant face recognition, we propose a self training solution of alternatively generating pseudo-labels and end-to-end re-training of the base CNN model. 
\medskip

Adapting a model trained on a source domain to a target domain where there is no labeled training data is referred as `unsupervised domain adaptation' (UDA). Feature space alignment \cite{ijcv14,fg18,icpr18}  and domain adversarial learning \cite{ssspdan,iccv17}  methods are common UDA face recognition practises. Feature space alignment aims to minimize the distance between domains in the feature space by learning transformations from source to target features. Domain adversarial learning transfers input domain in to target domain images without instance-level correspondence between domains, by using adversarial loss. Instead of minimizing the distance of the domains or confusing the domain discriminator using adversarial training, self training  generates  a  unified  feature  space  for  both  source  and target domains.
\medskip

This paper proposes a Self-Supervised Domain Learning (SSDL) scheme of iterative self learning where one begins with easy samples and adaptively progress towards the challenging.  Self-training is an underdeveloped research discipline in the context of face recognition. Co-training and adaptive facial model generation have similarities to self-training in that both approaches use operational data to build and update models. These adaptive facial models employ single or multiple classifiers like support vector machines, nearest neighbour classifiers, etc., that are updated based on face tracking results \cite{SSNaive2,SSNaive,ssNaive1}. Instead of training the shallow classifiers on top of deep features generated by the base CNN, SSDL directly remodels the base CNN. Hence, rather than learning compact and complicated decision boundaries, SSDL directly optimizes the feature space for more coherent boundary learning.  
\medskip  

Domain specific deep networks can be generated by two approaches: (1) joint-training on both source and target data, (2) pre-train on source data and fine-tune on target data. While both approaches achieve the goal of unified feature space for source and target, former has the risk of small domain-specific data being overpowered by large scale source data. In particular, while joint-training learns the global and task-specific discrepancies, re-training focuses only on target-specific discriminations and hence adapts better to the target. SSDL performs iterative fine-tuning where the two feature spaces are gradually and incrementally aligned by bringing the source closer to target in each iteration. SSDL also benefits from the advantage of having the network penalised by two different loss functions with varying strengths. In particular, classification based loss functions such as softmax loss is more related to tasks such as identity classification \cite{finetuneiccv17}, where as metric leaning approaches such as triplet loss directly reflect what is expected in face verification tasks \cite{FaceNet}. Following the work of \cite{VGGFace}, we use classification based loss function for primary training and metric learning based loss function for SSDL training to yield the advantages of both types of training.
\medskip

In summary, this paper focuses on the challenging problem of deep domain learning for face recognition aiming to dynamically adapt to different domains without labelled supervision. Our main contributions are as follows.

\begin{itemize}
  \item Building on a generic deep network model trained on labelled source data, we formulate SSDL as a latent variable loss minimization problem, by generating and learning on pseudo labels. 
  \item To generate pseudo labels on critical samples, we use a self-paced curriculum learning approach where the process begins with most confident samples and incrementally advance towards challenging samples through alternate  learning and predicting.
  \item To exploit the limited target train data generated on pseudo labels, we propose a semi-hard negative mining approach built on all positive samples.
  \item We evaluate the proposed approach on four different face recognition benchmarks designed to address different challenges in face recognition. We achieve significant levels of improvement over the baseline CNN while maintaining competitive performance with state-of-the art in all four benchmarks.
\end{itemize}

\section{Related Work} 

\subsection{Domain Adaptation for Face Recognition}

Over the recent years, domain adaptation for face recognition has been studied under three categories: supervised: labelled data are available in the target domain, semi-supervised: labelled and unlabelled data are available in the target domain and unsupervised: no labelled data in the target domain. Supervised approaches use labelled domain data to fine-tune the networks trained on source data. Semi-supervised approaches use a common/mediator database which is labelled and contains media from both source and target domains \cite{WEN201845}. Unsupervised domain learning approaches generally involve adversarial training \cite{iccv17,ssspdan}, transfer subspace \cite{icpr18,ijcv14}, dictionary learning \cite{7270280} and etc., where domain discrepancy is analysed using criteria such as  such as empirical maximum mean, mutual information, low-rank constraints and so on \cite{fg18}. This paper presents a self-supervised domain adaptation scheme which falls under unsupervised category since no labelled data is available in the target domain, but inherits from supervised approaches, since we perform fine-tuning on target domain data. 

\subsection{Self Training Based Domain Adaptation}

Self training strategy has been used in computer vision applications such object detection [33], semantic segmentation \cite{ECCV18} monocular face tracking \cite{ssCVPR} and face tracking \cite{Zhang2019} and clustering \cite{SSIAM}. Tang et al. \cite{tang}, Yoon et al. \cite{ssCVPR} and Zhang et al  \cite{Zhang2019} exploit temporal cues and constraints in generating train data.  Zhang et al. mines triplets where positive pairs are formed using small trajectories for each shot. Since a single shot spans across multiple frames, trajectories can be generated by calculating the intersection over union of bounding boxes on adjacent frames. Negative pairs are mined using two faces that appear in a single frame. This approach targets multi-face videos and is not applicable on face image sets or singleton videos.  Recently, Sharma et al. \cite{SSIAM} used the first and last k samples of a ranked list based on euclidean distance as strong candidates for positive and negative pairs. Preliminary fine-tuning on confident predictions, is applicable on tasks such as face tracking and clustering of videos, where tracking and clustering is guided by video specific spatio-temporal constraints. Nonetheless, `in-the-wild'/unconstrained face recognition requires a more profound learning and adaptation.
\medskip

Zou et al. \cite{ECCV18} (semantic segmentation) and Tang et al. \cite{tang} (object detection) used self-paced adaptation by learning via pseudo-labels in an easy-to-hard way. Zou et al. employed softmax cross-entropy loss to guide the self-training process. Our method is iterative and uses constraints that govern the number of examples to use in each iteration, which is similar to self-paced learning in \cite{ECCV18}. However, our method is different from   \cite{tang} and \cite{ECCV18}, in that we use triplet based training instead of class based training. In contrast to class based training, triplet loss allows more control over training process by enabling constrained triplet mining. By generating appropriate constraints, critical triplets can be generated to enhance the discriminative power to the level desired in fine-tuning. Moreover, Chu et al.  \cite{finetune} identifies 20 images per class as a threshold for deciding to fine-tune or freeze the first n layers when softmax loss is used. Developing on the said rule of thumb, one can safely assume that the expected domain learning can be achieved by via class based loss minimization on video face recognition where each motion spans across multiple frames and hence generates more than 20 faces per class. However, set based face recognition and image based recognition stipulates training on smaller class sizes.
\medskip

\section{Self-Supervised Domain Learning} 

\subsection{Preliminaries}

The purpose of penalizing a network with triplet loss is to ensure that an image $x^a_i$ (anchor) of a specific person is closer to all other images $x^p_i$ (positive) of the same person, than it is to any image $x^n_i$ (negative) of
any other person. For that, FaceNet \cite{FaceNet} defines the following two triplet constraints, equation \ref{cons1} being the violate constraint.
\medskip

\begin{equation}
|| x^a_i - x^p_i ||_2^2 + \alpha < || x^a_i - x^n_i ||_2^2  
\label{cons1}
\end{equation}

\begin{equation} 
|| x^a_i - x^p_i ||_2^2 < || x^a_i - x^n_i ||_2^2
\label{cons2}
\end{equation}

Selecting challenging triplets is crucial for faster convergence. VGGFace \cite{VGGFace} modifies the equation \ref{cons2} for a more challenging triplet mining approach. The equation \ref{cons2} ensures that the positive distance is less than the corresponding negative distance. The VGGFace equation states that additionally, the positive distance can be equal to negative distance. In doing so, the constraint in equation \ref{cons2} is replaced with the following.

\begin{equation} 
|| x^a_i - x^p_i ||_2^2 <= || x^a_i - x^n_i ||_2^2
\label{vggCons}
\end{equation}

The associated loss function is as follows.

\begin{equation} 
L = max\{0, || x^a_i - x^p_i ||_2^2 - || x^a_i - x^n_i ||_2^2 + \alpha\}
\label{mtl}
\end{equation}

\subsection{SSDL Design Considerations}

\begin{figure}
\centering
\includegraphics[height=5cm]{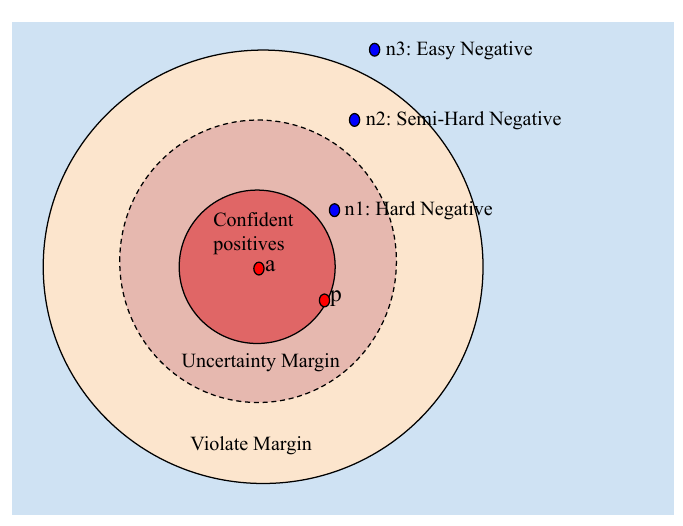}
\caption{Margins in SSDL. a: anchor, p: positive, n1,n2,n3: negatives}
\label{fig:tripletMargins}
\end{figure}

As exemplified in figure \ref{fig:tripletMargins}, the informative level of triplets vary depending on the distance levels between positives and negatives. Hence, constraints are introduced to filter the less informative. Conventional triplet loss uses a single constraint with a violate margin $\alpha$. While a single constraint setting is convenient on labelled data, informative triplet mining from noisy pseudo labels is more challenging and require additional constraints to filter uncertain samples. This paper proposes an uncertainty margin which discriminates strong predictions from ambivalent. While uncertainty margin also removes the hard negatives, it can be considered a reasonable trade-off between possible errors introduced by uncertain predictions and information learnt from hard samples.
\medskip

Given a train image set of size N, the number of possible triplets is $O(N)^3$. Regardless of the large number of candidate triplets, gradients are produced by the triplets that violate the triplet constraint. Hence, the violate margin $\alpha$ is a key factor in triplet mining. Intuitively, setting a higher value to $ \alpha$ can result in large number of easy triplets which lead to slow convergence and even convergence to local optima when a mini-batch is dominated by less informative triplets. Since SSDL is additionally constrained by a margin of uncertainty, setting a lower $\alpha$ can significantly reduce the number of triplets contributing to gradient learning and could ultimately result in over-fitting. Hence, we aim to generate a sufficient set of triplets by exploiting the prospective samples via definitive semi-hard triplet mining, as opposed to conventional random sampling. Additionally, we maintain lower learning rates to avoid learning a sub-optimal set of weights.
\medskip

The desirable feature of self-paced learning is to generate pseudo labels to easy samples in the initial iteration and to lean the hard samples in the iterations that follow. To excercise the process at its best, we aim to mine increasingly informative triplets by setting a decremental margin $\alpha$ which is reduced in each iteration. The simultaneous increase in the number of confident samples increases the number of triplets. This process, enables the last iterations to focus on the confusing examples and result in a more discriminative model.
\medskip

\begin{figure*}[htbp]
\centering

\pgfdeclarelayer{background}
\pgfdeclarelayer{foreground}
\pgfsetlayers{background,main,foreground}
\tikzstyle{conv}=[draw,thick,minimum width=3cm, fill=blue!10, ,rotate=90]  
\tikzstyle{loss}=[draw,thick , fill=yellow!30, font=\large,rotate=90]  

\tikzstyle{tbox}=[draw,thick,minimum height=0.9cm, text width=8em, fill=blue!10 ]

\resizebox {.8\linewidth}{!}{
\begin{tikzpicture}

    \node (source) {\includegraphics[height=2 cm]{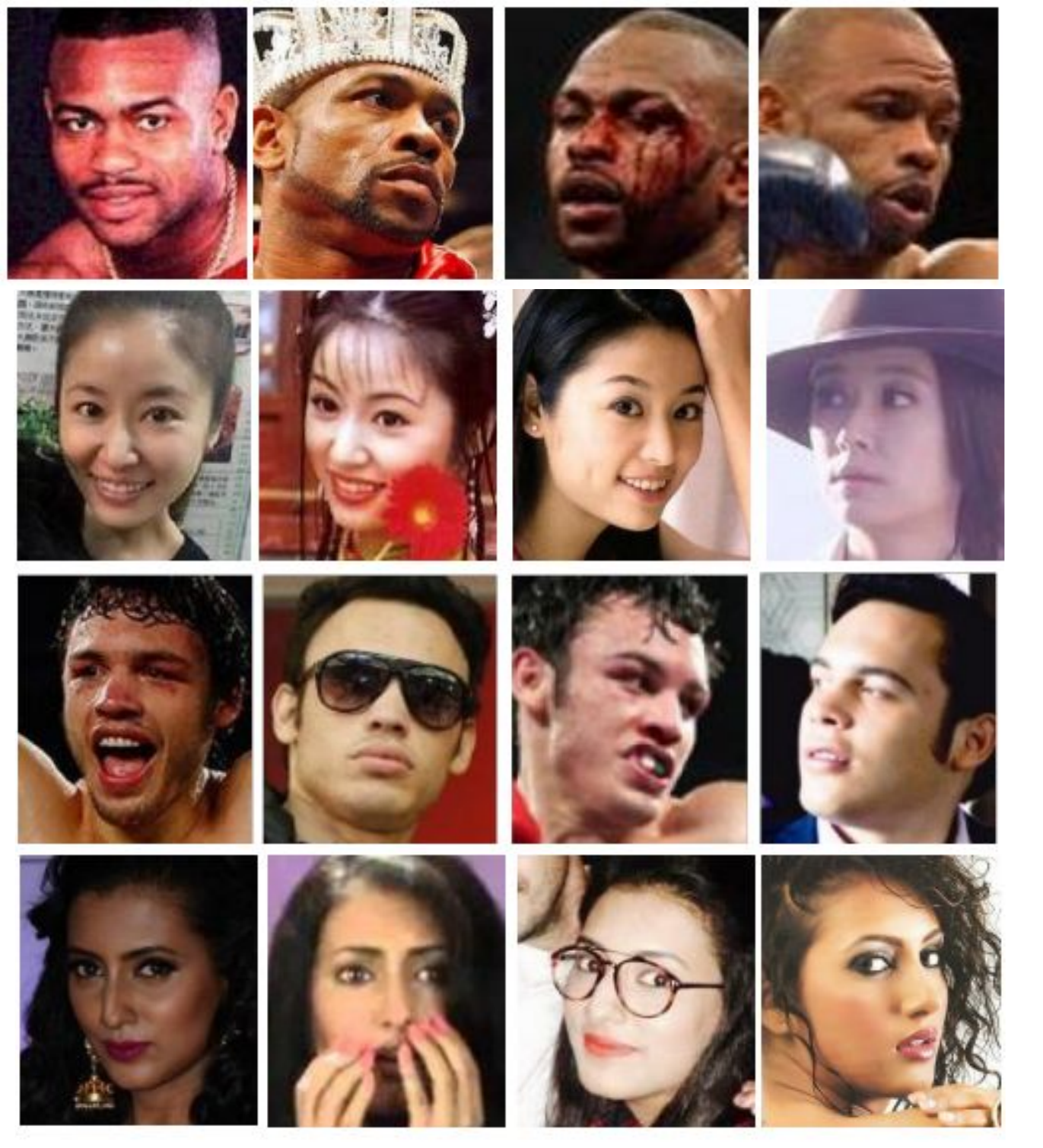}}; 
    
    \path   (source.south)+(0,-3) node (c_1_1) [conv,minimum height=0.8cm] {CNN};
    \path   (c_1_1.east)+(0,-3.3) node (l1) [ ] {(1) Pre-Training}; 
    \path   (c_1_1.east)+(0.7,1.1) node (l2) [] {\textit{Transfer Weights $\xrightarrow{}$}}; 
    \path   (c_1_1.east)+(0,4.1) node (l3) [ ] {Labelled Train Data}; 
    \path   (c_1_1.east)+(0,3.8) node (l4) [ ] {(Source Domain)}; 
    \path   (c_1_1.east)+(0.8,-1.5) node (c_2_1) [loss ] {Softmax Loss}; 
    \draw[red,thick, fill=red!5, 
      decoration={markings, mark=at position 0.125 with { \arrow[scale=1.5]{>}}},
      decoration={markings, mark=at position 0.400 with {\arrow[scale=1.5]{>}}},
      decoration={markings, mark=at position 0.650 with {\arrow[scale=1.5]{>}}}, 
      decoration={markings, mark=at position 0.875 with {\arrow[scale=1.5]{>}}}, 
      postaction={decorate}
      ] (c_2_1.east)+(5,1.6) circle (2.9cm);
    \draw[black,thick, fill=green!5 , 
      decoration={markings, mark=at position 0.125 with { \arrow[scale=1.5]{>}}},
      decoration={markings, mark=at position 0.400 with {\arrow[scale=1.5]{>}}},
      decoration={markings, mark=at position 0.650 with {\arrow[scale=1.5]{>}}}, 
      decoration={markings, mark=at position 0.875 with {\arrow[scale=1.5]{>}}}, 
      postaction={decorate} ] (c_2_1.east)+(5,1.2) circle (2.5cm);
    
     \draw [decorate,decoration={brace,mirror, amplitude=8pt},xshift=6pt,yshift=0pt]
(c_1_1.north east) -- (c_1_1.north west) node [black,midway,xshift=-0.6cm]
{};
    
    \path   (c_2_1.east)+(2.3,1) node (c_5) [conv,minimum height=0.8cm] {CNN}; 
    \path   (c_5.east)+(0.8,-1.5) node (c_2) [loss ] {Triplet Loss}; 
    \path   (c_2_1.east)+(5,-1.2) node (domain) [ ] {\includegraphics[height=2 cm]{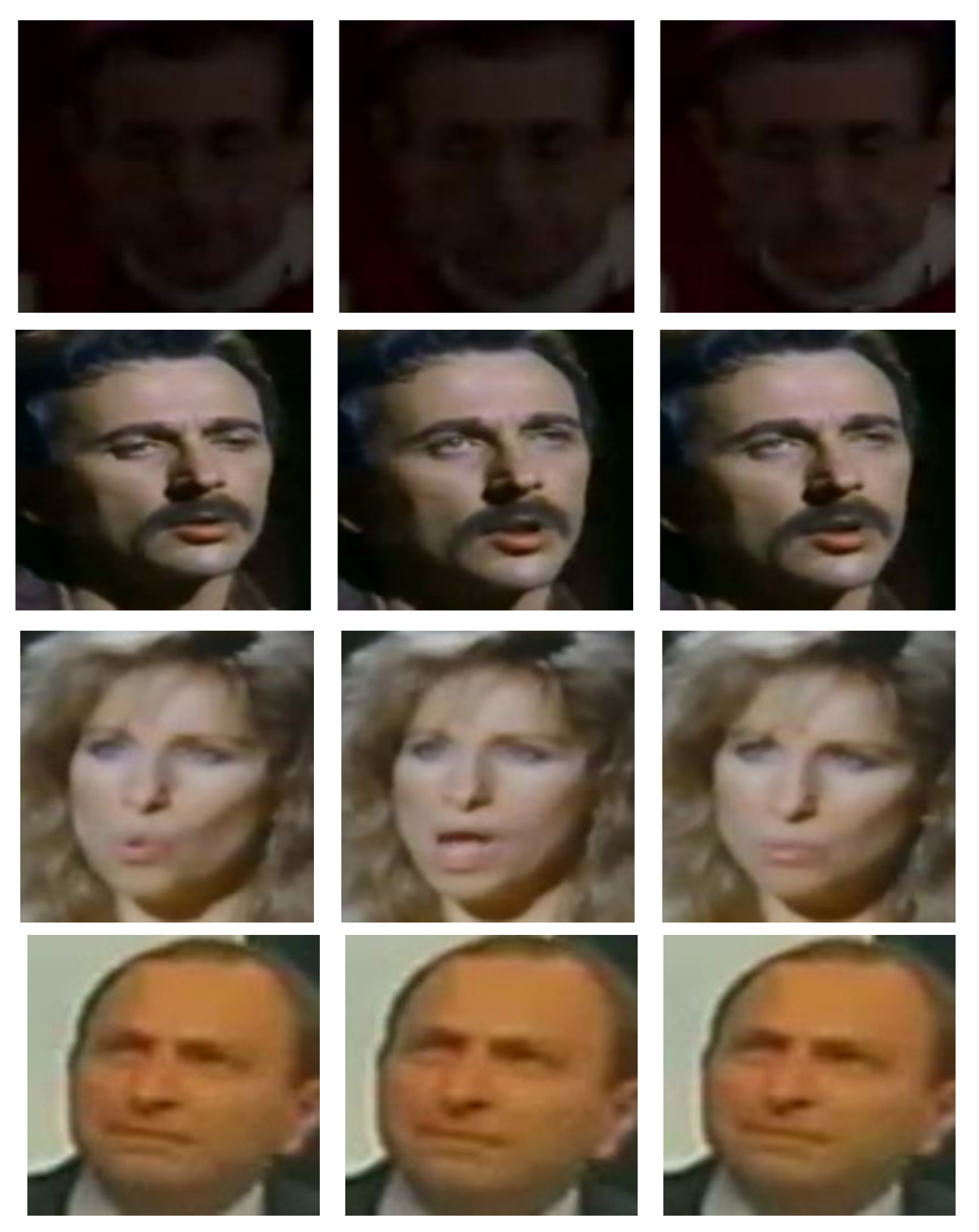}};
    \path   (domain.south)+(0,-0.2) node (l5) [ ] {Unlabelled Test Data}; 
    \path   (domain.south)+(0,-0.6) node (l6) [ ] {(Target Domain)}; 
    \path   (domain.south)+(0,3.6) node (l7) [ ] {(2) SSDL cycle}; 
    
    \path   (domain.east)+(1.5,2.6) node (clusters) [ ] {\includegraphics[height=1.5 cm]{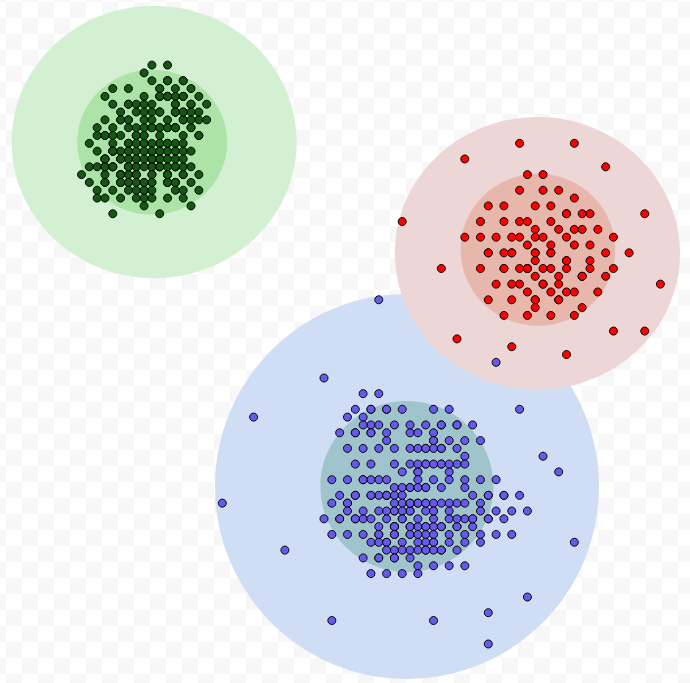}};    
    \path   (clusters.east)+(0.9,0) node (l8) [ ] {Salient Clusters}; 
    \path   (domain.east)+(-1.3,5.2) node (triplets) [ ] {\includegraphics[height=0.8 cm]{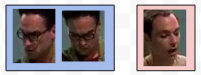}};   
    \path   (triplets.north)+(0,0.3) node (l8) [ ] {Triplets}; 
    \path   (triplets.south)+(-0.75,0) node (l8) [ ] {a};
    \path   (triplets.south)+(-0.15,0 ) node (l9) [ ] {p};
    \path   (triplets.south)+(0.65,0 ) node (l9) [ ] {n};
    \path   (domain.east)+(3.5,0) node (recog) [tbox] {(3) Recognition};

    \path [draw, ->] (source.south)+(0,0) -- node [] { } (c_1_1);  
    \path [draw, ->] (c_2_1.east)+(0.3,-1.2) -- node [] { } (domain); 
    \path [draw, ->] (domain.east)+(0,0) -- node [] { } (recog);      
    \draw[->, thick, black!40,line width=1.5mm]  (c_1_1.north)+(-0.35,0) -- ([xshift=-23pt, yshift=23pt]c_1_1.east) node[above left] {} -- (c_5);

\end{tikzpicture}}
\caption{The workflow of SSDL. a: anchor, p: positive, n:negative. Labelled train data samples are from VGGface2 \cite{VGGFace2} database, unlabelled test data samples are from YTF face recognition benchmark \cite{ytf}}
\label{fig:overview}
\end{figure*}
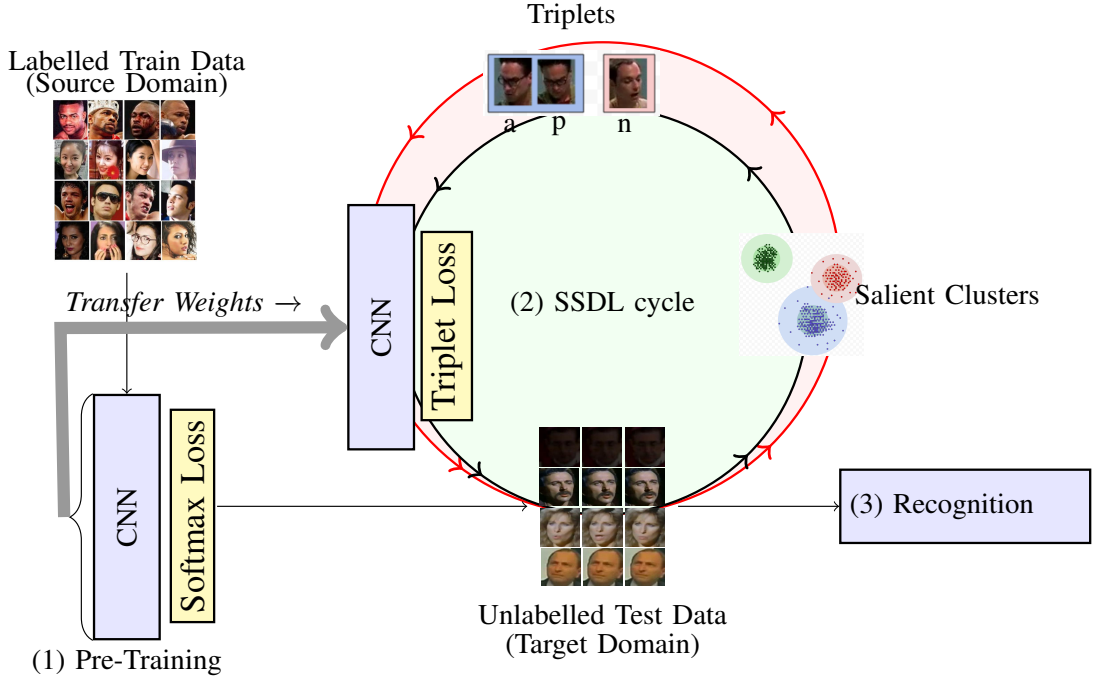

\subsection{SSDL Workflow}

As illustrated in figure \ref{fig:overview}, the SSDL begins with pre-training the model in source domain and is followed by domain specific training. In the pre-training phase, the deep CNN model is trained using the softmax loss function as optimization objective. Next, we transfer the weights from the initial training to initialize the the model in SSDL training cycle. The initialized model is used on the unlabelled data in the target domain to establish a set of salient clusters. The clusters are then used as the basis for generating pseudo labels on most confident predictions. We update the model weights by training on the pseudo labelled data, using triplet loss as optimization objective.
\medskip

Given a face detection $d$, the deep  network   maps  the  complex  high-dimensional  image information  into a  n-dimensional  proprietary feature  vector $\phi(d) \epsilon R^n$. The  generated  feature  vectors  can  be  interpreted  as  points in a  fixed-dimensional space where the euclidean distance between two points is analogous to the level of similarity between the two corresponding faces. Conventional face verification process employs a threshold $\beta$ such that if the euclidean distance between two faces is less than $\beta$, they are of the same individual. Building on this convention, we set margin of uncertainty $\gamma$ such that if the euclidean distance between two faces is less than $\beta - \gamma$, they are concluded as of the same individual with higher degree of certainty (i.e strong positives). Similarly, if the euclidean distance between two faces is greater than $\beta + \gamma$, they are concluded as of different individuals with higher degree of certainty (i.e strong negatives).
\medskip
 
\subsubsection{Salient Clustering} 

For clustering, we use a geometric approach, based on the assumption that each cluster $c_i$ is analogous to a sphere in the feature space with center $ctr(c_i)$ and radius R. Any data point that has distance more than R from the cluster center is a point outside the cluster sphere. We define the cluster radius $R = \beta/2 - \gamma/2$ and implement a margin based confident clustering algorithm, so that the following inequality holds for the elements in a single cluster $c_i$,
\medskip 

\begin{align}
|| \phi(x) - ctr(c_i) ||_2^2 < \beta/2 - \gamma/2
\label{clus}
\end{align}

where, $ctr()$, calculates the cluster center as follows. Given cluster $c_i$ with face detections $\{d^1, d^2, ...,d^n\}$,

\begin{align}
ctr(c_i) = mean(\phi(d^i), axis = 0), \forall { d^i \epsilon c_i} 
\label{ctr}
\end{align}

The clustering process is shown in algorithm \ref{Algo} where $D_f$ denotes the detections at frame $f$, $d_j$ the $j_{th}$ detection at that frame (for face images and singleton frames, $D_f = \{d\}$), C set of clusters and F the number of frames/images in the sequence.
\medskip

\begin{algorithm}
	\caption{Margin based confident clustering}
    \hspace*{\algorithmicindent} \textbf{Input:} $D = \{D_0, D_1, ..., D_{F-1}\}$ \\ \hspace*{\algorithmicindent} \hspace*{\algorithmicindent} \hspace*{\algorithmicindent} \hspace*{\algorithmicindent}  $ =
\{\{d_0, d_1, ..., d_{N_1}\}, \{d_0, d_1, ..., d_{N_2}\}, ...,  \\ \hspace*{\algorithmicindent} \hspace*{\algorithmicindent} \hspace*{\algorithmicindent} \hspace*{\algorithmicindent} \hspace*{\algorithmicindent} \hspace*{\algorithmicindent} \hspace*{\algorithmicindent}  \{d_0, d_1, ..., d_{N_{F-1}}\}\}$\\ 
    \hspace*{\algorithmicindent} \textbf{Initialize:} $C = \{\}$, 
    
	\begin{algorithmic}[1] 
	   
    	\FOR {f = 0 to F}
    	\FOR {$c_i$ $\epsilon$ $C$}
        	\STATE $d_{best} = d_j$ where $ min(||\phi(d_j) - ctr(c_i)||_2^2), d_j$ $\epsilon$ $D_f$ 
        	\IF{inequality \ref{clus} holds for $x=d_{best}$}
        	\STATE $c_i$ $\leftarrow$ $d_{best}$ 
        	\STATE update cluster center from equation \ref{ctr}
        	\STATE remove $d_{best}$ from $D_f$
        	\ENDIF
		\ENDFOR
		
    	\FOR {$d_j$ $\epsilon$ $D_f$}
    	\STATE initiate a new cluster with $d_j$
		\ENDFOR
		\ENDFOR
        \end{algorithmic} 
	\label{Algo}
\end{algorithm}

\subsubsection{Triplet Mining} 

Clustering process assigns pseudo labels to the unlabelled data such that all elements within the cluster has the same label. Positive samples are generated based on the generated pseudo labels. Next, we use the uncertainty margin $\gamma$ in the triplet constraints to mine negative samples. In doing so we use the violate constraint in equation \ref{cons1} and replace the obey constraint in equation \ref{cons2} with the following.

\begin{equation} 
|| x^a_i - x^p_i ||_2^2 + \gamma < || x^a_i - x^n_i ||_2^2
\label{cons2ssdl}
\end{equation}

Hence, the uncertainty margin based triplet loss is formulated as follows.

\begin{equation} 
L = max\{\gamma, || x^a_i - x^p_i ||_2^2 - || x^a_i - x^n_i ||_2^2 + \alpha\}
\label{mtl1}
\end{equation}

We formulate the problem of triplet mining as a definitive algorithm which exploits the limited triplets by  performing a semi-hard triplet mining approach. SSDL is constrained by the uncertainty margin as opposed to train data with accurate labels. During a given epoch, the proposed triplet mining algorithm generates the easiest negative for each anchor, positive pair subjected to constraints \ref{cons1} and \ref{cons2ssdl}. The proposed approach is presented in algorithm \ref{tripletAlgo} where $\Phi$ denotes the set of deep face descriptors ($\phi(x)$) with corresponding pseudo labels ($c_i$) and $T$ is the set of triplets. In terms of deep neural networks, an epoch refers to one cycle through the full training dataset. If the training data is fed to the network in multiple epochs with each epoch in a different pattern, one can hope for a better generalization at unseen test data. We provide triplets of increasing information levels in each epoch. In particular, the first epoch uses the easiest negatives with in the constraints, and the second epoch uses the second easiest negatives and so on.

\begin{algorithm}
	\caption{Triplet Collection}
    \hspace*{\algorithmicindent} \textbf{Input:} $\Phi = [\{\phi(x_1), c_1\}, \{\phi(x_2), c_2\}, ..., \{\phi(x_{n}), c_n\}]$, \\
    \hspace*{\algorithmicindent}\hspace*{\algorithmicindent}\hspace*{\algorithmicindent}\hspace*{\algorithmicindent}\hspace*{\algorithmicindent} \text{where }$\phi(x) \epsilon R^d$, \\
    \hspace*{\algorithmicindent}\hspace*{\algorithmicindent}\hspace*{\algorithmicindent}\hspace*{\algorithmicindent}\hspace*{\algorithmicindent}n= \text{number of clustered faces} \\  
    \hspace*{\algorithmicindent}\hspace*{\algorithmicindent}\hspace*{\algorithmicindent}\hspace*{\algorithmicindent}\hspace*{\algorithmicindent} $c_i=$ \text{cluster label} \\  
    \hspace*{\algorithmicindent}\hspace*{\algorithmicindent}\hspace*{\algorithmicindent}\hspace*{\algorithmicindent}\hspace*{\algorithmicindent} epoch = \text{ Training epoch} \\  
    \hspace*{\algorithmicindent} \textbf{Initialize: } $T = \{\}$, 
    
	\begin{algorithmic}[1] 
	    
    	\FOR {each $\{a,p\} \text{ where } c_a = c_p$ }
    	    \STATE  all\_negs = $\{\phi(x_a), c_i\}, \forall{i \epsilon \{1,2, ..., N\}} \text{ where } a!=i$  
    	\STATE $neg\_dists=[]$ 
    	    
    	\FOR {$n$ $\epsilon$ all\_negs }
        	\IF{constraint in equation \ref{cons1} violates AND constraint in equation \ref{cons2ssdl} holds}
        	
        	\STATE $neg\_dists \leftarrow DIST(a,n)$ 
        	\ENDIF
        	
		\ENDFOR 
        	
		\STATE Inversely sort $neg\_dists$
		\STATE $n \leftarrow (epoch+1)^{th}$ element in $(neg\_dists)$
		\STATE $T \leftarrow {a,p,n}$
        	
		\ENDFOR
		
        \end{algorithmic} 
	\label{tripletAlgo}
\end{algorithm}

\subsubsection{Self-Paced Domain Learning}

The proposed scheme of self-paced domain learning is comprised of two iterations, the first being a Domain-Blind (DB) triplet mining process and the second, a Domain-Aware (DA) process. Each iteration has three steps, step 1: salient clustering, step 2: triplet mining and step 3: model updating. Intuitively, domain blind triplet collection selects the most confident samples. Once the model is updated, and better adapted to target domain, we repeat the clustering process expecting a more solid clustering result. Assuming a more concrete clustering result, the margin of uncertainty $\gamma$ is reduced in domain aware triplet collection. As the uncertainty level lowers, the number of strong triplets rises, enabling a more sensitive filter which retains the most informative samples. The advanced filtering is achieved by reducing the triplet violate margin $\alpha$.

\subsection{Implementation Details}

 We use an implementation of the MTCNN architecture described in \cite{MTCNN} for face detection.  To leverage the recent advent of deep CNN architectures, we use Inception ResNet V1 network discussed in \cite{InceptionV4}, trained with softmax loss. The network is trained with VGGFace2 dataset \cite{VGGFace2} which contains 3.31 million face images.
 \medskip

We use the raw video frames or images without any supporting labelled information for SSDL training. To minimize the impact of random false detections during the clustering stage, clusters with length less than a pre-defined size (we used 5 in our experiments) are considered unstable clusters and are discarded. We derive $\beta$ by evaluating the pre-trained model on the conventional LFW \cite{lfw} face verification benchmark and by obtaining the threshold corresponding to the highest verification accuracy. We use the parameters $\alpha = 0.2$ which is empirically derived by the authors of FaceNet \cite{FaceNet}. We set $\gamma=0.1$.  In domain aware triplet mining we reduce the values of $\alpha$ and $\gamma$ to $\alpha = 0.1$ and $\gamma=0.05$.
 \medskip

Chu et al. sets \cite{finetune} the learning rate to 0.2 times that of random initialization in fine-tuning. Given the limited number of triplets available for training after the filtering of uncertain predictions, we set the learning rate as 0.03 times that of conventional model training.
\medskip
 
We formulate both video face recognition and set based recognition as a set based recognition problem and perform feature aggregation to assign a single representation to each set of faces. For a more reliable representation, we heavily weigh the stable face appearances using the normalised detection scores, as follows.
 
 \begin{equation}
    \phi(S) = \sum_k w_{s_i}\phi(s_i), \forall s_i \epsilon S 
\end{equation}

where $\phi(S)$ is the set level representation of face image set $S=\{s_1, s_2, ...s_n\}$ of n faces. $w_{s_i}$ is the weight corresponding to the face $s_i$.
\medskip

\subsection{Benchmarks} 
  
\subsubsection{YouTube Celebrities Face Recognition Dataset}

The YouTube Celebrities (YTC) video dataset consists of 1,910 video sequences of 47 celebrities from YouTube.  There are large variations of pose, illumination, and expression on face videos in this dataset. Moreover, the quality of face videos is very poor because most videos are of high compression rate. The experiment setting is the same as \cite{tyctrans,MMDML}. Five fold cross validation was carried out with three video sequences per subject for training and six for testing in each fold. 
\medskip

\setlength{\tabcolsep}{4pt}
\begin{table}
\begin{center}
\caption{Classification Rates (\%) on the
YouTube Celebrities Dataset.}
\label{ytc}
\begin{tabular}{lc}
\hline\noalign{\smallskip}
System & Classification Accuracy (\%)\\
\noalign{\smallskip}
\hline
\noalign{\smallskip} 
SANP (TPAMI 12) \cite{tyctrans}  & 65.60 \\
MMDML (CVPR 15) \cite{MMDML} & 78.5  \\
DRM-PWV  (TPAMI 15)\cite{pami} & 72.55  \\
Fast FR (ICCVW 17) \cite{fastAcc} & 72.1\\
GJRNP (IVC 17) \cite{YANG201747} & 81.3 \\
ClusterFace (ICPR20) \cite{ClusterFace} & 91.06 \\
\hline
Baseline model & 88.43\\ 
SSDL on DB triplets & \textcolor{red}{\textbf{94.75}} \\
SSDL on DA triplets & \textcolor{red}{\textbf{94.75}} \\ 
\hline
\end{tabular}
\end{center}
\end{table}
\setlength{\tabcolsep}{1.4pt}
\setlength{\tabcolsep}{4pt}

\subsubsection{The YouTube Faces Dataset}

The YouTube Faces dataset (YTF) \cite{ytf} is larger than YTC video dataset and contains 3425 videos of 2595 different subjects. The lengths of videos vary from 48 to 6070 frames and the average length is 181.3 frames per video.
The dataset is split into 10 folds, and each fold consists of 250 positive (intra-subject) pairs and 250 negative (inter-subject) pairs. The benchmark targets the pair matching problem/face verification under two protocols: (1) restricted : the pairs are provided and (2) unrestricted : the pairs can be generated as per user's preference. We follow the restricted  protocol to test our method and report the verification accuracy.
\medskip

\subsubsection{IARPA Janus Benchmark A }
The IJB-A Dataset (IJB-A) \cite{ijb_a} contains 5712 images and 2085 videos of 500 subjects. The average numbers of images and videos per subject are 11.4 images and 4.2 videos. The images are manually aligned as opposed to the general practise of using a commodity  face  detector. The manual alignment process preserves challenging variations such as pose, occlusion, illumination and etc., that are generally filtered out with automated detection. The dataset is a collection of media in the wild which contains both images and videos. Hence, this dataset includes multi-target domains. Therefore we use the ClusterFace \cite{ClusterFace} approach for image set based recognition, on top of our features in experiments on IJB-A.  
\medskip

\begin{table}
\begin{center}
\caption{Verification Performance on the YouTube Faces Database (Restricted Protocol)}
\label{ytf}
\begin{tabular}{l c c }
\hline\noalign{\smallskip}
System & Accuracy (\%)\\
\noalign{\smallskip}
\hline
\noalign{\smallskip}
EigenPEP (ACCV 14) \cite{EigenPEP} & 84.80 \\
DeepFace-single (CVPR 15) \cite{DeepFace} & 91.40  \\
DeepID2+ (CVPR 15) \cite{DeepIdTWOPLUS} & 93.20  \\
FaceNet (CVPR 15) \cite{FaceNet} & 95.12 \\ 
CenterLoss (ECCV 16) \cite{CenterLoss} & 94.9  \\
NAN (CVPR 16) \cite{NAN} & \textcolor{red}{\textbf{95.72}}  \\
TBE-CNN (TPAMI 18) \cite{TBE} & 94.96  \\
\hline
Baseline model & 93.92 \\ 
SSDL on DB triplets & 95.01 \\
SSDL on DA triplets & \textcolor{blue}{\textbf{95.66}} \\ 
\hline
\end{tabular}
\end{center}
\end{table}

\subsubsection{Cox Face Database}

The COX Face database \cite{COX} comprises of 1,000 still images and 3,000 surveillance like videos of 1,000 subjects. Each walking individual was captured simultaneously from three cameras at different locations. The V2V identification protocol requires individual identification across the three cameras. In addition, all the subjects are of a single ethnicity, which makes it complementary to the other benchmarks. Hence Cox Face provides a platform to evaluate the ability of the proposed scheme to self-formulate and learn intra-ethnic disciminative features. 
\medskip

\subsection{Comparison With the Baseline Model}

Tables \ref{ytc}, \ref{ytf}, \ref{ijba}, \ref{cox} reports the evaluation results on YTC, YTF, IJB-A and Cox Face databases. It is clear that the proposed approach achieves considerable improvements over the baseline CNN on all four databases. In particular, it reports a 6.32\% increase in classification accuracy on YTC, 1.74\% increase in verification accuracy in YTF database, 9.91\%, 1.87\% increase in true acceptance rate at  0.01, 0.1 false positive rates in IJB-A and an average 9.33\% increase in identification accuracy on Cox Face database. The experimental evaluation shows two results: (1) the SSDL generalises well on different domains, (2) re-training on the confident samples (domain blind triplets) certainly does improve recognition, but the critical samples (domain-aware triplets) contribute more to performance enhancement.
\medskip

\begin{figure}
\centering
\includegraphics[height=5cm]{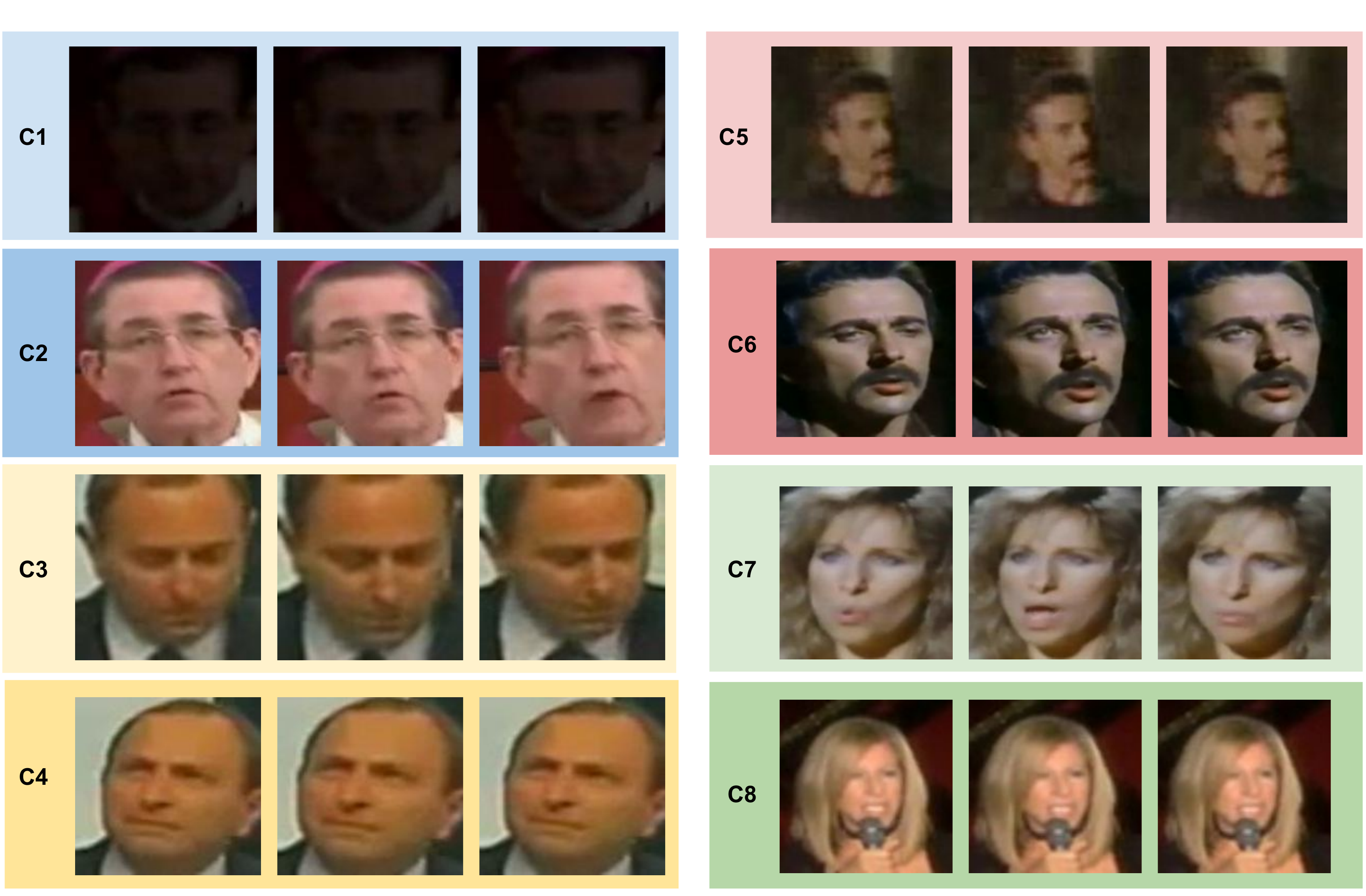}
\caption{Left: Clusters that were left unmerged in domain blind clustering and were correctly merged during domain aware clustering. According to ground truth, C1, and C2 belong to one individual, C3, C4 belong to another individual. The two pairs were correctly merged during the second clustering cycle. Right: Clusters that were left unmerged during both clustering cycles. According to ground truth, C5, C6 belong to a single individual and C7, C8 belong to a different single individual. None-the-less both pairs were left unmerged during the two clustering cycles. Data source: YTF benchmark \cite{ytf}}
\label{fig:ClusteringErr}
\end{figure}

One of the key goals in iterative learning is to learn increasingly challenging samples in each iteration. Such learning can be achieved by an increasingly accurate clustering result in each iteration. Figure \ref{fig:ClusteringErr} shows samples of clustering errors encountered during the training cycles. (C1, C2) and (C3, C4) are clusters that were not merged during first clustering cycle but were correctly merged during the second, whereas (C5, C6) and (C7, C8) were left unmerged during both clustering cycles. Triplets across such unmerged clusters of the same individual are expected to be filtered out by the uncertainty margin. 
\medskip

\setlength{\tabcolsep}{4pt}
\begin{table}
\begin{center}
\caption{Performance evaluation for verification on IJB-A benchmark. The true accept rates (TAR) vs. false positive rates (FAR) are reported.}
\label{ijba}
\begin{tabular}{lccc}
\hline\noalign{\smallskip}
System &  FAR=0.001 &  FAR=0.01 & FAR=0.1\\
\noalign{\smallskip}
\hline
\noalign{\smallskip}
 
Triplet Similarity \cite{tripSimilarity} & 59.0 & 79.0 & 94.5 \\
Multi-pose (WACV16) \cite{DeepMultiPose} & - & 78.7 & 91.1 \\
Triplet Emb (BTAS16) \cite{tripletEmb} & 81.3 & 91 & 96.4 \\
FastSearch (TPAMI17) \cite{FastSearch} & 51.0& 72.9 & 89.3 \\
Joint Bayesian (WACV16) \cite{jointBaysian} & - & 83.8 & 96.7 \\
PAM (CVPR16) \cite{poseawarecvpr} & 65.2 & 82.6 & - \\
NAN (CVPR16) \cite{NAN} & \textcolor{blue}{\textbf{88.1}} & 94.1 & 97.8  \\
Template (FG17)  \cite{template} & 83.6 &  93.9 & 97.9\\
DR GAN (CVPR17) \cite{DRGAN}& 53.9 & 77.4 & - \\ 
Contrastive (ECCV18) \cite{contrastiveCNN} & 63.91 & 84.01 & 95.31 \\
ClusterFace (ICPR20) \cite{ClusterFace} &  86.60 & \textcolor{blue}{\textbf{94.23}} & \textcolor{red}{\textbf{98.30}} \\ 
\hline
Baseline model & 63.84 & 84.82 & 96.24 \\ 
SSDL on DB triplets & 77.98 & 92.09 & 97.83 \\
SSDL on DA triplets & \textcolor{red}{\textbf{88.77}} & \textcolor{red}{\textbf{94.73}} & \textcolor{blue}{\textbf{98.11}}\\  
 
\hline
\end{tabular}
\end{center}
\end{table}
\setlength{\tabcolsep}{1.4pt}

\subsection{Comparison with the State-Of-The-Art}

Furthermore, our proposed method is compared with a some of the important state-of-the-art  face recognition systems. All our experiments assume zero knowledge on the target domain and hence does not use the provided train data to fine-tune the models. The highest (denoted in red) and the second highest (denoted in blue) results are highlighted in each benchmark results. It is clear that a plain network without any data augmentation techniques or using multiple ensembles and under single crop experiments, achieves better or competitive results on all occasions.

\begin{table}
\begin{center}
\caption{Identification Rates (\%) under the V2V Setting for Different Methods on the COX Face Database}
\label{cox}
\begin{tabular}{lllllll}
\hline\noalign{\smallskip}
System & V2-V1 & V3-V1 & V3-V2 & V1-V2 & V1-V3 & V2-V3 \\
\noalign{\smallskip}
\hline
\noalign{\smallskip}
PSCL \cite{COX} & 57.70 & 73.17 & 67.70 & 62.77 & 78.26 & 68.91 \\
LERM (CVPR 14) & 65.94 & 78.24& 70.67 & 64.44 & 80.53 & 72.96 \\
HERML-GMM (PR 15) \cite{HUANG20153113}& 95.10 & 96.30 & 94.20 & 92.30 & \textcolor{blue}{\textbf{95.40}} & 94.50 \\ 
TBE-CNN (TPAMI 18) \cite{TBE} & \textcolor{red}{\textbf{98.07}} &  \textcolor{red}{\textbf{98.16}} & \textcolor{red}{\textbf{97.93}} & \textcolor{red}{\textbf{97.20}} & \textcolor{red}{\textbf{99.30}} & \textcolor{red}{\textbf{99.33}} \\
\hline
Baseline model & 91.07 & 87.49 & 84.79 & 91.84 & 82.27 & 79.44 \\
SSDL on DB triplets & 90.14 & 92.31  & 94.44 & 87.48 & 87.0 & 91016\\
SSDL on DA triplets &  \textcolor{blue}{\textbf{96.41}} & \textcolor{blue}{\textbf{96.4}} & \textcolor{blue}{\textbf{96.01}} & \textcolor{blue}{\textbf{95.82}} & 93.07& \textcolor{blue}{\textbf{95.17}}
 \\ 
\hline
\end{tabular}
\end{center}
\end{table}
\setlength{\tabcolsep}{1.4pt}

\section{Conclusions} 

Domain adaptation for face recognition is of paramount interest for preserving model reliability across different domains. Accordingly, we have presented self-supervised domain learning scheme, which adapts by self learning when labelled supervision is unavailable. Alternatively, samples from the target domain are automatically collected to supervise model re-training. We formulate the sample selection problem in an easy-to-hard way where we begin with the most confident samples and progress to confusing samples. We have tested our proposals on four different benchmarks. The experimental evaluations show  that proposed approach  lead to considerably better or competitive performance compared to many current state of the art face recognition systems.

\section*{Acknowledgment}

The research activities leading to this publication has been partly funded by the European Union Horizon 2020 Research and Innovation program under grant agreement No. 786629 (MAGNETO RIA project).






\bibliographystyle{IEEEtran}
\bibliography{main}

\begin{thebibliography}{10}
\providecommand{\url}[1]{#1}
\csname url@samestyle\endcsname
\providecommand{\newblock}{\relax}
\providecommand{\bibinfo}[2]{#2}
\providecommand{\BIBentrySTDinterwordspacing}{\spaceskip=0pt\relax}
\providecommand{\BIBentryALTinterwordstretchfactor}{4}
\providecommand{\BIBentryALTinterwordspacing}{\spaceskip=\fontdimen2\font plus
\BIBentryALTinterwordstretchfactor\fontdimen3\font minus
  \fontdimen4\font\relax}
\providecommand{\BIBforeignlanguage}[2]{{%
\expandafter\ifx\csname l@#1\endcsname\relax
\typeout{** WARNING: IEEEtran.bst: No hyphenation pattern has been}%
\typeout{** loaded for the language `#1'. Using the pattern for}%
\typeout{** the default language instead.}%
\else
\language=\csname l@#1\endcsname
\fi
#2}}
\providecommand{\BIBdecl}{\relax}
\BIBdecl

\bibitem{FaceNet}
F.~{Schroff}, D.~{Kalenichenko}, and J.~{Philbin}, ``Facenet: A unified
  embedding for face recognition and clustering,'' in \emph{2015 IEEE
  Conference on Computer Vision and Pattern Recognition (CVPR)}, Jun. 2015, pp.
  815--823.

\bibitem{DeepIdTWOPLUS}
Y.~Sun, X.~Wang, and X.~Tang, ``Deeply learned face representations are sparse,
  selective, and robust,'' in \emph{2015 IEEE Conference on Computer Vision and
  Pattern Recognition (CVPR)}, 2015, pp. 2892--2900.

\bibitem{DeepFace}
Y.~{Taigman}, M.~{Yang}, M.~{Ranzato}, and L.~{Wolf}, ``Deepface: Closing the
  gap to human-level performance in face verification,'' in \emph{2014 IEEE
  Conference on Computer Vision and Pattern Recognition}, Jun. 2014, pp.
  1701--1708.

\bibitem{FaceSurvSAM}
\BIBentryALTinterwordspacing
S.~{Wickrama Arachchilage} and E.~{Izquierdo}, ``Deep-learned faces: a
  survey,'' \emph{EURASIP Journal on Image and Video Processing}, 2020.
  [Online]. Available: \url{https://doi.org/10.1186/s13640-020-00510-w}
\BIBentrySTDinterwordspacing

\bibitem{ijb_a}
B.~Klare, B.~Klein, E.~Taborsky, A.~Blanton, J.~Cheney, K.~E. Allen,
  P.~Grother, A.~Mah, M.~Burge, and A.~K. Jain, ``Pushing the frontiers of
  unconstrained face detection and recognition: Iarpa janus benchmark a,'' in
  \emph{{2015 IEEE Conference on Computer Vision and Pattern Recognition
  (CVPR)}}, 2015, pp. 1931--1939.

\bibitem{NIPS2014_5347}
\BIBentryALTinterwordspacing
J.~Yosinski, J.~Clune, Y.~Bengio, and H.~Lipson, ``How transferable are
  features in deep neural networks?'' in \emph{Advances in Neural Information
  Processing Systems 27}, Z.~Ghahramani, M.~Welling, C.~Cortes, N.~D. Lawrence,
  and K.~Q. Weinberger, Eds.\hskip 1em plus 0.5em minus 0.4em\relax Curran
  Associates, Inc., 2014, pp. 3320--3328. [Online]. Available:
  \url{http://papers.nips.cc/paper/5347-how-transferable-are-features-in-deep-neural-networks.pdf}
\BIBentrySTDinterwordspacing

\bibitem{ijcv14}
M.~Kan, J.~Wu, S.~Shan, and X.~Chen, ``Domain adaptation for face recognition:
  Targetize source domain bridged by common subspace,'' \emph{International
  Journal of Computer Vision}, vol. 109, 08 2014.

\bibitem{fg18}
Z.~{Luo}, J.~{Hu}, W.~{Deng}, and H.~{Shen}, ``Deep unsupervised domain
  adaptation for face recognition,'' in \emph{2018 13th IEEE International
  Conference on Automatic Face Gesture Recognition (FG 2018)}, May 2018, pp.
  453--457.

\bibitem{icpr18}
H.~{Yi}, Z.~{Xu}, Y.~{Wen}, and Z.~{Fan}, ``Multi-source domain adaptation for
  face recognition,'' in \emph{2018 24th International Conference on Pattern
  Recognition (ICPR)}, Aug 2018, pp. 1349--1354.

\bibitem{ssspdan}
S.~Hong, W.~Im, J.~Ryu, and H.~Yang, ``Sspp-dan: Deep domain adaptation network
  for face recognition with single sample per person,'' 02 2017.

\bibitem{iccv17}
K.~{Sohn}, S.~{Liu}, G.~{Zhong}, X.~{Yu}, M.~{Yang}, and M.~{Chandraker},
  ``Unsupervised domain adaptation for face recognition in unlabeled videos,''
  in \emph{2017 IEEE International Conference on Computer Vision (ICCV)}, Oct
  2017, pp. 5917--5925.

\bibitem{SSNaive2}
N.~El~Gayar, S.~A. Shaban, and S.~Hamdy, ``Face recognition with
  semi-supervised learning and multiple classifiers,'' in \emph{Proceedings of
  the 5th WSEAS International Conference on Computational Intelligence,
  Man-Machine Systems and Cybernetics}, ser. CIMMACS’06.\hskip 1em plus 0.5em
  minus 0.4em\relax Stevens Point, Wisconsin, USA: World Scientific and
  Engineering Academy and Society (WSEAS), 2006, p. 296–301.

\bibitem{SSNaive}
K.~{Lu}, Z.~{Ding}, J.~{Zhao}, and Y.~{Wu}, ``A novel semi-supervised face
  recognition for video,'' in \emph{2010 International Conference on
  Intelligent Control and Information Processing}, Aug 2010, pp. 313--316.

\bibitem{ssNaive1}
\BIBentryALTinterwordspacing
M.~D. la~Torre, E.~Granger, P.~V. Radtke, R.~Sabourin, and D.~O. Gorodnichy,
  ``Partially-supervised learning from facial trajectories for face recognition
  in video surveillance,'' \emph{Information Fusion}, vol.~24, pp. 31 -- 53,
  2015. [Online]. Available:
  \url{http://www.sciencedirect.com/science/article/pii/S1566253514000700}
\BIBentrySTDinterwordspacing

\bibitem{finetuneiccv17}
A.~{Hasnat}, J.~{Bohné}, J.~{Milgram}, S.~{Gentric}, and L.~{Chen},
  ``Deepvisage: Making face recognition simple yet with powerful generalization
  skills,'' in \emph{2017 IEEE International Conference on Computer Vision
  Workshops (ICCVW)}, 2017, pp. 1682--1691.

\bibitem{VGGFace}
\BIBentryALTinterwordspacing
O.~M. Parkhi, A.~Vedaldi, and A.~Zisserman, ``Deep face recognition,'' in
  \emph{Proceedings of the British Machine Vision Conference (BMVC)}.\hskip 1em
  plus 0.5em minus 0.4em\relax BMVA Press, Sept. 2015, pp. 41.1--41.12.
  [Online]. Available: \url{https://dx.doi.org/10.5244/C.29.41}
\BIBentrySTDinterwordspacing

\bibitem{WEN201845}
\BIBentryALTinterwordspacing
G.~Wen, H.~Chen, D.~Cai, and X.~He, ``Improving face recognition with domain
  adaptation,'' \emph{Neurocomputing}, vol. 287, pp. 45 -- 51, 2018. [Online].
  Available:
  \url{http://www.sciencedirect.com/science/article/pii/S0925231218301127}
\BIBentrySTDinterwordspacing

\bibitem{7270280}
Q.~{Qiu} and R.~{Chellappa}, ``Compositional dictionaries for domain adaptive
  face recognition,'' \emph{IEEE Transactions on Image Processing}, vol.~24,
  no.~12, pp. 5152--5165, Dec 2015.

\bibitem{ECCV18}
Y.~Zou, Z.~Yu, B.~V.~K. Vijaya~Kumar, and J.~Wang, ``Unsupervised domain
  adaptation for semantic segmentation via class-balanced self-training,'' in
  \emph{Computer Vision -- ECCV 2018}, V.~Ferrari, M.~Hebert, C.~Sminchisescu,
  and Y.~Weiss, Eds.\hskip 1em plus 0.5em minus 0.4em\relax Cham: Springer
  International Publishing, 2018, pp. 297--313.

\bibitem{ssCVPR}
J.~S. {Yoon}, T.~{Shiratori}, S.~{Yu}, and H.~S. {Park}, ``Self-supervised
  adaptation of high-fidelity face models for monocular performance tracking,''
  in \emph{2019 IEEE/CVF Conference on Computer Vision and Pattern Recognition
  (CVPR)}, June 2019, pp. 4596--4604.

\bibitem{Zhang2019}
\BIBentryALTinterwordspacing
S.~Zhang, J.-B. Huang, J.~Lim, Y.~Gong, J.~Wang, N.~Ahuja, and M.-H. Yang,
  ``Tracking persons-of-interest via unsupervised representation adaptation,''
  \emph{International Journal of Computer Vision}, Sep. 2019. [Online].
  Available: \url{https://doi.org/10.1007/s11263-019-01212-1}
\BIBentrySTDinterwordspacing

\bibitem{SSIAM}
V.~Sharma, M.~Tapaswi, M.~S. Sarfraz, and R.~Stiefelhagen, ``Self-supervised
  learning of face representations for video face clustering,'' in \emph{IEEE
  International Conference on Automatic Face and Gesture Recognition (FG
  2019)}, 2019.

\bibitem{tang}
\BIBentryALTinterwordspacing
K.~Tang, V.~Ramanathan, L.~Fei-fei, and D.~Koller, ``Shifting weights: Adapting
  object detectors from image to video,'' in \emph{Advances in Neural
  Information Processing Systems 25}, F.~Pereira, C.~J.~C. Burges, L.~Bottou,
  and K.~Q. Weinberger, Eds.\hskip 1em plus 0.5em minus 0.4em\relax Curran
  Associates, Inc., 2012, pp. 638--646. [Online]. Available:
  \url{http://papers.nips.cc/paper/4691-shifting-weights-adapting-object-detectors-from-image-to-video.pdf}
\BIBentrySTDinterwordspacing

\bibitem{finetune}
B.~Chu, V.~Madhavan, O.~Beijbom, J.~Hoffman, and T.~Darrell, ``Best practices
  for fine-tuning visual classifiers to new domains,'' in \emph{ECCV
  Workshops}, 2016.

\bibitem{VGGFace2}
Q.~Cao, L.~Shen, W.~Xie, O.~M. Parkhi, and A.~Zisserman, ``Vggface2: A dataset
  for recognising faces across pose and age,'' in \emph{International
  Conference on Automatic Face and Gesture Recognition (FG)}, 2018.

\bibitem{ytf}
L.~{Wolf}, T.~{Hassner}, and I.~{Maoz}, ``Face recognition in unconstrained
  videos with matched background similarity,'' in \emph{2011 IEEE Conference on
  Computer Vision and Pattern Recognition}, Jun. 2011, pp. 529--534.

\bibitem{MTCNN}
K.~{Zhang}, Z.~{Zhang}, Z.~{Li}, and Y.~{Qiao}, ``Joint face detection and
  alignment using multitask cascaded convolutional networks,'' \emph{IEEE
  Signal Processing Letters}, vol.~23, no.~10, pp. 1499--1503, Oct. 2016.

\bibitem{InceptionV4}
C.~Szegedy, S.~Ioffe, V.~Vanhoucke, and A.~A. Alemi, ``Inception-v4,
  inception-resnet and the impact of residual connections on learning,'' in
  \emph{Thirty-First AAAI Conference on Artificial Intelligence}, 2017.

\bibitem{lfw}
\BIBentryALTinterwordspacing
G.~B. Huang, M.~Mattar, T.~Berg, and E.~Learned-Miller, ``{Labeled Faces in the
  Wild: A Database for Studying Face Recognition in Unconstrained
  Environments},'' in \emph{{Workshop on Faces in 'Real-Life' Images:
  Detection, Alignment, and Recognition}}.\hskip 1em plus 0.5em minus
  0.4em\relax Marseille, France: {Erik Learned-Miller and Andras Ferencz and
  Fr{\'e}d{\'e}ric Jurie}, Oct. 2008. [Online]. Available:
  \url{https://hal.inria.fr/inria-00321923}
\BIBentrySTDinterwordspacing

\bibitem{tyctrans}
Y.~{Hu}, A.~S. {Mian}, and R.~{Owens}, ``Face recognition using sparse
  approximated nearest points between image sets,'' \emph{IEEE Transactions on
  Pattern Analysis and Machine Intelligence}, vol.~34, no.~10, pp. 1992--2004,
  Oct 2012.

\bibitem{MMDML}
J.~{Lu}, G.~{Wang}, W.~{Deng}, P.~{Moulin}, and J.~{Zhou}, ``Multi-manifold
  deep metric learning for image set classification,'' in \emph{2015 IEEE
  Conference on Computer Vision and Pattern Recognition (CVPR)}, June 2015, pp.
  1137--1145.

\bibitem{pami}
M.~{Hayat}, M.~{Bennamoun}, and S.~{An}, ``Deep reconstruction models for image
  set classification,'' \emph{IEEE Transactions on Pattern Analysis and Machine
  Intelligence}, vol.~37, no.~4, pp. 713--727, April 2015.

\bibitem{fastAcc}
H.~{Cevikalp} and H.~S. {Yavuz}, ``Fast and accurate face recognition with
  image sets,'' in \emph{2017 IEEE International Conference on Computer Vision
  Workshops (ICCVW)}, Oct 2017, pp. 1564--1572.

\bibitem{YANG201747}
\BIBentryALTinterwordspacing
M.~Yang, X.~Wang, W.~Liu, and L.~Shen, ``Joint regularized nearest points for
  image set based face recognition,'' \emph{Image and Vision Computing},
  vol.~58, pp. 47 -- 60, 2017. [Online]. Available:
  \url{http://www.sciencedirect.com/science/article/pii/S026288561630124X}
\BIBentrySTDinterwordspacing

\bibitem{ClusterFace}
S.~{Wickrama Arachchilage} and E.~{Izquierdo}, ``Clusterface: Joint clustering
  and classification for set-based face recognition,'' in \emph{25th
  International Conference on Pattern Recognition}, 2020.

\bibitem{EigenPEP}
H.~Li, G.~Hua, X.~Shen, Z.~Lin, and J.~Brandt, ``Eigen-pep for video face
  recognition,'' in \emph{Computer Vision -- ACCV 2014}, D.~Cremers, I.~Reid,
  H.~Saito, and M.-H. Yang, Eds.\hskip 1em plus 0.5em minus 0.4em\relax Cham:
  Springer International Publishing, 2015, pp. 17--33.

\bibitem{CenterLoss}
Y.~Wen, K.~Zhang, Z.~Li, and Y.~Qiao, ``A discriminative feature learning
  approach for deep face recognition,'' in \emph{Computer Vision -- ECCV 2016},
  B.~Leibe, J.~Matas, N.~Sebe, and M.~Welling, Eds.\hskip 1em plus 0.5em minus
  0.4em\relax Cham: Springer International Publishing, 2016, pp. 499--515.

\bibitem{NAN}
J.~Yang, P.~Ren, D.~Chen, F.~Wen, H.~Li, and G.~Hua, ``Neural aggregation
  network for video face recognition,'' \emph{2017 IEEE Conference on Computer
  Vision and Pattern Recognition (CVPR)}, pp. 5216--5225, 2016.

\bibitem{TBE}
C.~{Ding} and D.~{Tao}, ``Trunk-branch ensemble convolutional neural networks
  for video-based face recognition,'' \emph{IEEE Transactions on Pattern
  Analysis and Machine Intelligence}, vol.~40, no.~4, pp. 1002--1014, April
  2018.

\bibitem{COX}
Z.~{Huang}, S.~{Shan}, R.~{Wang}, H.~{Zhang}, S.~{Lao}, A.~{Kuerban}, and
  X.~{Chen}, ``A benchmark and comparative study of video-based face
  recognition on cox face database,'' \emph{IEEE Transactions on Image
  Processing}, vol.~24, no.~12, pp. 5967--5981, Dec 2015.

\bibitem{tripSimilarity}
\BIBentryALTinterwordspacing
S.~Sankaranarayanan, A.~Alavi, and R.~Chellappa, ``Triplet similarity embedding
  for face verification,'' \emph{CoRR}, vol. abs/1602.03418, 2016. [Online].
  Available: \url{http://arxiv.org/abs/1602.03418}
\BIBentrySTDinterwordspacing

\bibitem{DeepMultiPose}
W.~{AbdAlmageed}, Y.~{Wu}, S.~{Rawls}, S.~{Harel}, T.~{Hassner}, I.~{Masi},
  J.~{Choi}, J.~{Lekust}, J.~{Kim}, P.~{Natarajan}, R.~{Nevatia}, and
  G.~{Medioni}, ``Face recognition using deep multi-pose representations,'' in
  \emph{2016 IEEE Winter Conference on Applications of Computer Vision (WACV)},
  March 2016, pp. 1--9.

\bibitem{tripletEmb}
S.~{Sankaranarayanan}, A.~{Alavi}, C.~D. {Castillo}, and R.~{Chellappa},
  ``Triplet probabilistic embedding for face verification and clustering,'' in
  \emph{2016 IEEE 8th International Conference on Biometrics Theory,
  Applications and Systems (BTAS)}, Sep. 2016, pp. 1--8.

\bibitem{FastSearch}
D.~{Wang}, C.~{Otto}, and A.~K. {Jain}, ``Face search at scale,'' \emph{IEEE
  Transactions on Pattern Analysis and Machine Intelligence}, vol.~39, no.~6,
  pp. 1122--1136, June 2017.

\bibitem{jointBaysian}
J.~{Chen}, V.~M. {Patel}, and R.~{Chellappa}, ``Unconstrained face verification
  using deep cnn features,'' in \emph{2016 IEEE Winter Conference on
  Applications of Computer Vision (WACV)}, March 2016, pp. 1--9.

\bibitem{poseawarecvpr}
I.~{Masi}, S.~{Rawls}, G.~{Medioni}, and P.~{Natarajan}, ``Pose-aware face
  recognition in the wild,'' in \emph{2016 IEEE Conference on Computer Vision
  and Pattern Recognition (CVPR)}, June 2016, pp. 4838--4846.

\bibitem{template}
N.~Crosswhite, J.~Byrne, C.~Stauffer, O.~Parkhi, Q.~Cao, and A.~Zisserman,
  ``Template adaptation for face verification and identification,'' in
  \emph{2017 12th IEEE International Conference on Automatic Face \& Gesture
  Recognition (FG 2017)}, May 2017, pp. 1--8.

\bibitem{DRGAN}
L.~{Tran}, X.~{Yin}, and X.~{Liu}, ``Disentangled representation learning gan
  for pose-invariant face recognition,'' in \emph{2017 IEEE Conference on
  Computer Vision and Pattern Recognition (CVPR)}, July 2017, pp. 1283--1292.

\bibitem{contrastiveCNN}
C.~Han, S.~Shan, M.~Kan, S.~Wu, and X.~Chen, ``Face recognition with
  contrastive convolution,'' in \emph{Computer Vision -- ECCV 2018},
  V.~Ferrari, M.~Hebert, C.~Sminchisescu, and Y.~Weiss, Eds.\hskip 1em plus
  0.5em minus 0.4em\relax Cham: Springer International Publishing, 2018, pp.
  120--135.

\bibitem{HUANG20153113}
\BIBentryALTinterwordspacing
Z.~Huang, R.~Wang, S.~Shan, and X.~Chen, ``Face recognition on large-scale
  video in the wild with hybrid euclidean-and-riemannian metric learning,''
  \emph{Pattern Recognition}, vol.~48, no.~10, pp. 3113 -- 3124, 2015,
  discriminative Feature Learning from Big Data for Visual Recognition.
  [Online]. Available:
  \url{http://www.sciencedirect.com/science/article/pii/S0031320315001120}
\BIBentrySTDinterwordspacing

\end{thebibliography}
%



\end{document}